\def\year{2018}\relax
\documentclass[letterpaper]{article} 
\usepackage{aaai18}  
\usepackage{times}  
\usepackage{helvet}  
\usepackage{courier}  
\usepackage{url}  
\usepackage{graphicx}  
\usepackage{booktabs} 
\usepackage{mathtools}
\usepackage{amsfonts}
\begin{document}
%
\title{Movie Question Answering: \\Remembering the Textual Cues for Layered Visual Contents}
\author{Bo Wang, Youjiang Xu, Yahong Han\\
School of Computer Science and Technology,\\
Tianjin University, Tianjin, China\\
\{bwong, yjxu, yahong\}@tju.edu.cn\\
\And
Richang Hong\\
School of Computer and Information,\\
Hefei University of Technology, Hefei, China\\
hongrc.hfut@gmail.com\\
}

\maketitle
\begin{abstract}
Movies provide us with a mass of visual content as well as attracting stories. Existing methods have illustrated that understanding movie stories through only visual content is still a hard problem. In this paper, for answering questions about movies, we put forward a Layered Memory Network (LMN) that represents frame-level and clip-level movie content by the Static Word Memory module and the Dynamic Subtitle Memory module, respectively. Particularly, we firstly extract words and sentences from the training movie subtitles. Then the hierarchically formed movie representations, which are learned from LMN, not only encode the correspondence between words and visual content inside frames, but also encode the temporal alignment between sentences and frames inside movie clips. We also extend our LMN model into three variant frameworks to illustrate the good extendable capabilities. We conduct extensive experiments on the MovieQA dataset. With only visual content as inputs, LMN with frame-level representation obtains a large performance improvement. When incorporating subtitles into LMN to form the clip-level representation, we achieve the state-of-the-art performance on the online evaluation task of `Video+Subtitles'. The good performance successfully demonstrates that the proposed framework of LMN is effective and the hierarchically formed movie representations have good potential for the applications of movie question answering.
\end{abstract}

\section{Introduction}
Bridging the visual understanding and computer-human interaction is a challenging task in artificial intelligence. Though visual captioning \cite{li2017image,dmrmyangMM17,liu2017video,pan2016hierarchical,wang2012event} has shown to be promising in connecting the visual content to natural languages, it usually narrates the coarse semantic of visual content and lacks abilities of modeling different correlations among visual cues. Whereas visual question answering (VQA) \cite{malinowski2015ask,zhu2016visual7w,xiong2016dynamic} relies on the holistic scene understanding to find correct answers for different levels of visual understanding. Popular methods towards VQA aim to learn the co-occurrence of a particular combination of features extracted from images and questions, e.g., space embedding for images and words via Convolutional Neural Networks (CNNs) and Recurrent Neural Networks (RNNs) \cite{malinowski2015ask}. In order to accurately associate specific language elements with particular visual content, attention \cite{zhu2016visual7w} or dynamic memory \cite{xiong2016dynamic} mechanisms were proposed to improve the performance of VQA.

As videos could be taken as a spatio-temporal extension of images, video understanding requires a better representation to encode both the visual content of each frame and the temporal dependencies among successive video frames. Different from other videos, most movies have a specific background (e.g., action film and war film) as well as the shooting style (e.g., flashback). Thus understanding the story of a movie via only the visual content is really a challenging task. On the other hand, a movie always contains a standard subtitle which consists of the dialogues between actors. This offers a possibility to better understand the story of a movie and thus facilitate applications of automatically movie question answering.


\begin{figure*}
\centering
\includegraphics[width=0.95\linewidth]{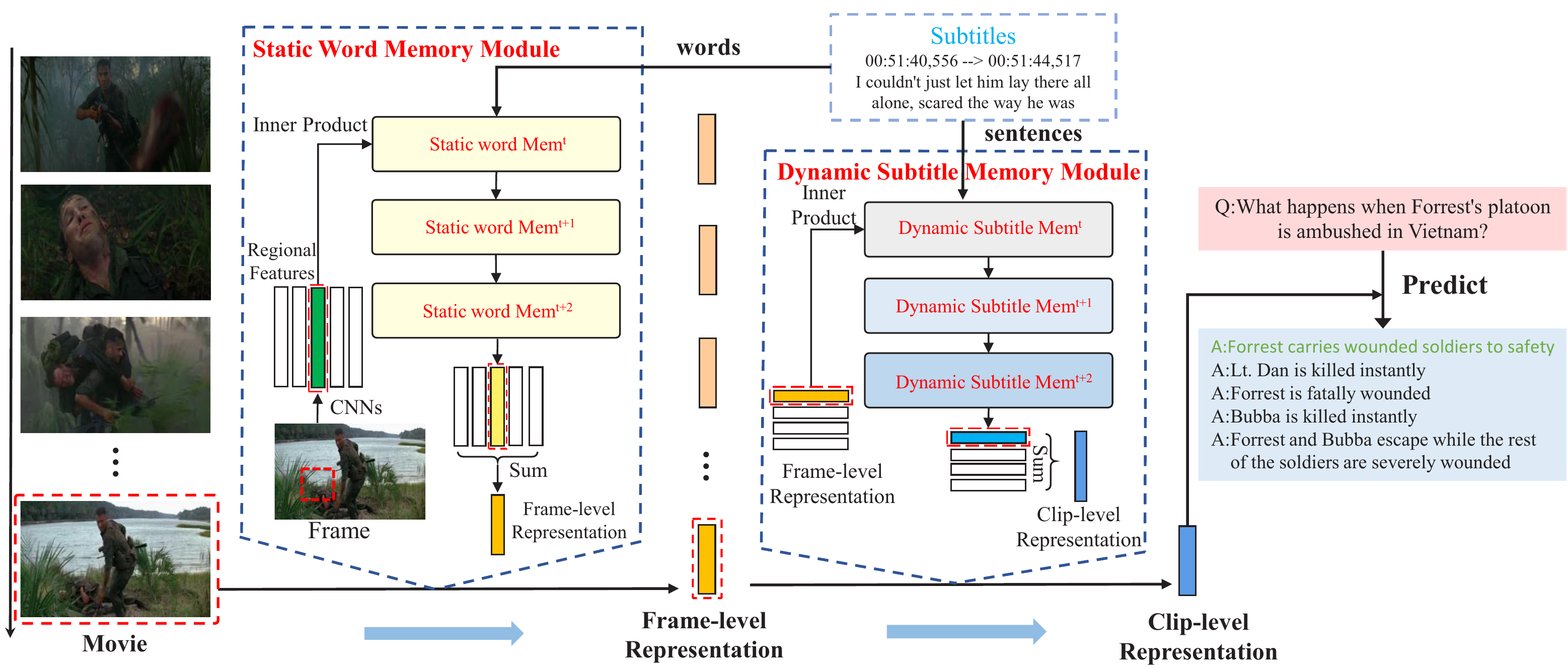}
\caption{The framework of our method. We firstly get frame-level representation by representing regional features of each movie frame with the Static Word Memory module. Then the frame-level representation is fed into the Dynamic Subtitle Memory module to obtain the final clip-level representation, which is applied to answer the questions.}
\label{fig:framework}
\end{figure*}

In this paper, we explore how to utilize movie clips and subtitles for movie question answering. We propose a Layered Memory Network (LMN) to learn a layered representation of movie content, i.e., frame-level and clip-level layers, by a Static Word Memory module and a Dynamic Subtitle Memory module, respectively. The Static Word Memory contains all the words information of the MovieQA dataset \cite{tapaswi2016movieqa} while the Dynamic Subtitle Memory contains all the subtitle information. The framework of our proposed method is illustrated in Fig. \ref{fig:framework}. Firstly, we get the frame-level representation by representing regional features of each movie frame with Static Word Memory module. Secondly, the generated frame-level representation is fed into the Dynamic Subtitle Memory module to obtain the final clip-level representation. Thus, the hierarchically formed movie representations, i.e., the layered representation learned by LMN from frame and clip-level, not only encode the correspondence between words and visual content inside frames, but also encode the temporal alignment between sentences and frames inside movie clips.

The main contributions of this paper are summarized as follows. (1) We propose a Layered Memory Network which can utilize visual and text information. The model can represent movie content with more semantic information in both frame-level and clip-level. (2) We propose three extended frameworks base on our LMN, which can remove the irrelevant information from massive external texts and can improve the reasoning ability of the LMN model. (3) The LMN method shows good performance of movie question answering on the MovieQA dataset. We obtain the state-of-the-art performance on the online evaluation\footnote{\url{http://movieqa.cs.toronto.edu/leaderboard/#table-movie}} task of `Video+Subtitles' for movie question answering.

\section{Related Work}\label{Related Work}

\subsection{Image Question Answering} Methods towards image question answering are mainly categorized into joint embedding \cite{ren2015image,ma2016learning}, attention mechanism \cite{zhu2016visual7w,yang2016stacked,shih2016look,xu2016ask}, and incorporating external knowledge \cite{nie2013beyond,wang2016fvqa,wu2016ask,zhu2017cvpr} etc. The joint embedding methods target to learn the co-occurrence of a particular combination of features extracted from images and questions, whereas attention mechanism is utilized to accurately associate specific language elements with particular visual content. As knowledge-based reasoning is effective in traditional question answering system, experiments have shown that appropriately utilizing external knowledge could improve the performance of image question answering, such as storing the vectorized facts into a dynamic memory network \cite{xiong2016dynamic} or connecting the CNN-extracted visual concepts to the node on a knowledge graph \cite{wang2015explicit}.



\subsection{Video Question Answering} As videos could be taken as spatio-temporal extensions of images, how to incorporate the temporal cues into the video representation and associate it to certain textual cues in questions is crucial to the video question answering \cite{zhu2015uncovering,zeng2017leveraging}. As videos are more complex than images, datasets construction to boost the research of video question answering is a challenge task, such as TGIF-QA \cite{jang-CVPR-2017}, MarioQA \cite{mun2017marioqa}, the `fill-in-the-blank' \cite{zhu2015uncovering} and the large-scale video question answering dataset without manual annotations \cite{zeng2017leveraging}. Recently, Tapaswi et al. \cite{tapaswi2016movieqa} proposed a MovieQA dataset with multiple sources for movie question answering, which successfully attracted interesting work, such as video-story learning \cite{ijcai2017-280} and multi-modal movie question answering \cite{na2017movie}. Though the Deep Embedded Memory Networks (DEMN) \cite{ijcai2017-280} could reconstruct stories from a joint scene-dialogue video stream, it is not trained in an end-to-end way. The Read-Write Memory Network (RWMN) \cite{na2017movie} utilizes multi-layered CNNs to capture and store the sequential information of movies into the memory, so as to help answer movie questions. As a specific application of video question answering, movie question answering requires both accurate visual information and high-level semantic to infer the answer. Subtitles accompanied movie clips may implicitly narrate the story cues temporally aligned with sequential frames, or even convey relationships among roles and objects.


\section{The Proposed Method}\label{The Proposed Method}

In this section, we introduce the proposed method of Layered Memory Network (LMN), which utilizes both movie contents and subtitles to answer the movie question. The input of the LMN is a sequence of frame-wise feature maps $\{I_1, I_2, \dots, I_T\}$ and question $u$. The output is the correct answer which is predicted by the LMN. In the following subsections, we firstly introduce how the LMN represent the visual content on both frame-level and clip-level in word space and sentence space. As the proposed LMN is a basic framework for movie question answering, we then propose three variants to demonstrate the extendable capability of our framework. 

\subsection{Represent Movie Frames with Static Word Memory Module}
\label{Represent Movie Frames with Static Word Memory}
\begin{figure}
\centering
\includegraphics[width=0.95\linewidth]{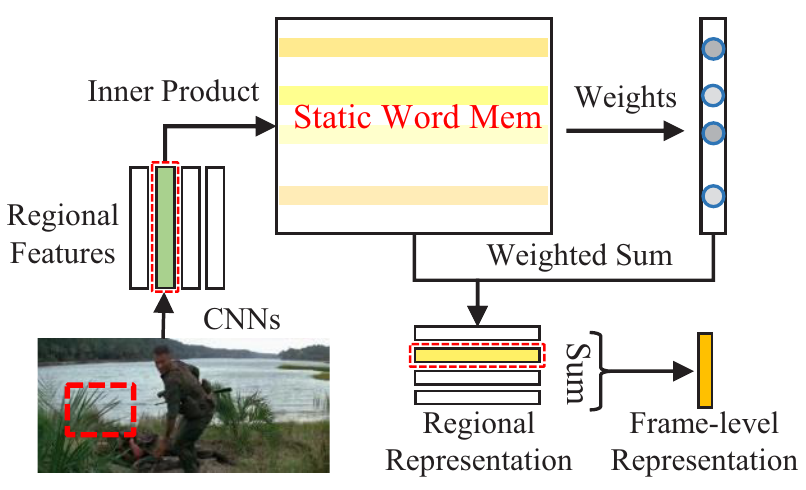}
\caption{The framework of Static Word Memory module. }
\label{fig:add_1}
\end{figure}

The main objective of this module is to get a semantic representation of specific region in the movie frames through Static Word Memory. The framework is shown in Fig. \ref{fig:add_1}. Suppose we have a Static Word Memory of $W_e \in \mathbb{R}^{|\mathcal{V}|\times d}$, which can be seen as a word embedding matrix with a vocabulary size $|\mathcal{V}|$ that maps words into a continuous vector of $d$-dimension. The Static Word Memory can be learned by skip-gram model \cite{mikolov2013efficient}. There is a sequence of frame-wise feature maps $\{I_1, I_2, \dots, I_T\}$, which are extracted from a convolutional layer of CNNs, and each of which has the shape of $C \times H \times W$, where $C, H, W$ denote the channel, height, and width of feature maps, respectively. Thus we can obtain $H\times W$ regional features $\{l_{ij} \in \mathbb{R}^C\}$ and $i \in \{1, 2, \dots, T\}, j \in \{1, 2, \dots, H \times W\}$. Different from the joint embedding methods which directly map the regional features into a common space, we represent the regional features with the Static Word Memory. We first map our regional features into the word space with a projection $W_l \in \mathbb{R}^{d \times C}$ by
$v_{ij} = W_ll_{ij}$, where the ${v_{ij}}$ can be seen as the $j$-th projected regional feature of $i$-th movie frame. Then we utilize an inner product to compute the similarity between the projected regional features and words of the Static Word Memory. The formulation is defined as:
\begin{align}\label{weight-of-regions}
  \alpha_{ijk} = v_{ij}^Tw_k,
\end{align}
where $w_k$ denotes the $k$-th row vector of Static Word Memory $W_e$. Both $v_{ij}$ and $w_k$ are first scaled to have unit norm, thus the $\alpha_{ijk}$ is equivalent to the cosine similarity. Then the regional feature $v_{ij}$ can be replaced by a weighted sum over all words of the memory:
\begin{align}\label{weighted-sum-of-regions}
  v_{ij} \coloneqq \sum_{k=1}^{|\mathcal{V}|}\alpha_{ijk}w_k,
\end{align}
where $|\mathcal{V}|$ denotes the size of our Static Word Memory. The $\coloneqq$ represents the `update' operation. Thus the frame-level representation can be computed by:

\begin{align}\label{frame-level-rep}
  v_i = \sum_{j=1}^{H\times W}v_{ij}.
\end{align}
Because each regional feature is a weighted sum over the whole words vectors, the $v_i$ can be seen as a semantic representation of $i$-th movie frame. This procedure is similar as the attention mechanism on the regions of an image. However, our model attends to the memory of the words for each regional features.

The proposed Static Word Memory has two properties: (1) Static Word Memory could be taken as a word embedding matrix. We can utilize the memory to map word into a continuous vector. (2) Static Word Memory could be utilized to represent the regional features of movie frames.

\subsection{Represent Movie Clips with Dynamic Subtitle Memory Module}
\label{Represent Movie Clips with Dynamic Subtitle Memory}
The main objective of this module is to get a semantic representation of the specific frame in movie clips through Dynamic Subtitle Memory. As a movie contains not only visual content but also subtitles, we put forward a Dynamic Subtitle Memory module to represent movie clips with movie subtitles. Suppose the movie frames have been represented as frame-level representation ${v_i}$ which is the output of Eq. \eqref{frame-level-rep}. Then the subtitles $\{s_1,s_2,\dots,s_N\}$, which are first embedded by the Static Word Memory, are utilized as Dynamic Subtitle Memory. We have a similar representing procedure:

\begin{align}
  \beta_i^n &= v_i^Ts_n, \label{weight-of-frame}\\
  v_i &\coloneqq \sum_{n=1}^{N} \beta_i^ns_n, \label{weighted-sum-of-frames}\\
  v &= \sum_{i=1}^{T}v_i, \label{clip-level-ave-rep}
\end{align}
where $\beta_i^n$ denotes the similarity of $n$-th subtitle corresponding to the $i$-th frame, and the Eq. \eqref{weighted-sum-of-frames} represents the $i$-th frame representation is replaced by weighted summing over all subtitles. The clip-level representation $v$ can be obtained by summing over the frame-level representation $\{v_i\}$. For the task of movie question answering with video as the only inputs, the frame-level representation $\{v_i\}$ in Eq. \eqref{clip-level-ave-rep} is the output of Eq. \eqref{frame-level-rep}. As a result, our clip-level representation is transformed from word space into sentence space and thus can obtain much semantic information. Then we use the method from \cite{tapaswi2016movieqa} to answer the movie questions with open-ended answers as follows:

\begin{align}
  a &= softmax((v+u)^Tg), \label{video-prediction}
\end{align}
where $u$ denotes the question vector, $g = \{g_1, g_2,\dots,g_5\}$ and $g_h$ denote the $h$-th answer. Both question and answers are embedded by the Static Word Memory $W_e$. Note that the Static Word Memory is shared for representing the regional features of movie frames and words of sentences. However, the Subtitle Memory is different for each movie as different movies may have unique subtitles. Thus our Word Memory is static but our Subtitle Memory is dynamic. And both Static Word Memory model and Dynamic Subtitle Memory model discussed above only have one memory layer.

In summary, the Layered Memory Network has following advantages: (1) Instead of learning the joint embedding matrices, we directly replace the regional features and frame-level features by the Static Word Memory and the Dynamic Subtitle Memory, respectively. The layered frame-level and clip-level representations contain richer semantic information and achieve good performance on movie question answering, which will be discussed in detail in our experiments. (2) Benefiting from exploiting accuracy words and subtitles to represent each regional and frame-level features, our method obtains a good property of supporting accurate reasoning while answering a question. For example, we can find out the most relative subtitle of each movie frame. Some of our results are shown in Fig.\ref{fig:example_1} and Fig. \ref{fig:example_2}. (3) Our method is a basic framework which has a good extendable capability. We will introduce some extended frameworks based on our method in the following section. 

\begin{figure}
\centering
\includegraphics[width=0.95\linewidth]{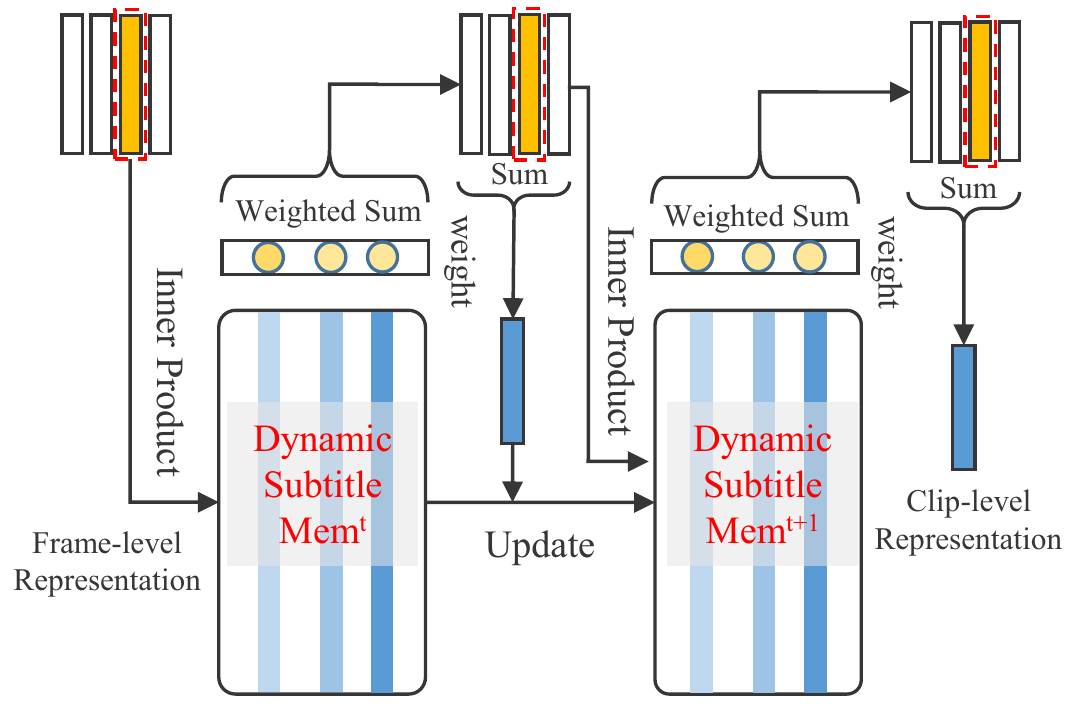}
\caption{The framework of Dynamic Subtitle Memory module with update mechanism. The circle represents the weight computed by the inner product between frame-level representation and the subtitle.}
\label{fig:extend_2}
\end{figure}

\subsection{The Extended Frameworks}

\subsubsection{Multiple Hops in Static Word Memory.}
As described above, we only use a single Static Word Memory, which means the region representation is obtained only by one mapping process. In this case, the region representation may contain much irrelevant information and lack some key content to answer the question. So according to the multiple hops mechanism \cite{sukhbaatar2015end} in memory network, we consider using multiple hops in Static Word Memory to get a better frame-level semantic representation. Different from the memory network which uses the sum of the output ${o_k}$ and the input ${u_k}$ of layer ${k}$ as the input of next layer. We only use the regions' semantic representation, which are the output of the ${t}$-th Static Word Memory, as the input of the ${(t+1)}$-th Static Word Memory. So Eq. \eqref{weight-of-regions}-\eqref{weighted-sum-of-regions} can be replaced as:
\begin{align}\label{Multiple-layers}
  \alpha_{ijk}^{t+1} &= {(v_{ij}^{t})}^Tw_{k}^{t+1},\\
  v_{ij}^{t+1} &= \sum_{k=1}^{|\mathcal{V}|}\alpha_{ijk}^{t+1}w_{k}^{t+1},
\end{align}
The regional representation ${v_{ij}^{t}}$ is the output of the $t$-th Static Word Memory. The $w_{k}^{t+1}$ is the $k$-th row of the ${(t+1)}$-th Static Word Memory. The ${\alpha_{ijk}^{t+1}}$ denotes the cosine similarity between ${v_{ij}^{t}}$ and $w_{k}^{t+1}$. Note that, the Static Word Memory we use to do multiple hops are the same, $w_{k}^{t+1} = w_{k}^{t}$. And the Static Word Memory consists of words in all movie subtitles. Thus the `Static' has two meanings here: (1) The Word Memory $W_e$ is shared by all movies in the MovieQA dataset. (2) The Word Memory $W_e$ remains the same during the multiple hops in Static Word Memory.

\subsubsection{Update Mechanism of Dynamic Subtitle Memory.} The subtitle contains complete dialogues in the movie. However, the movie clips only contains certain segments of the whole movie. So even we utilize the weighted sum over the original subtitles to form the clip-level representation, there will still contain a lot of irrelative information in the clip-level representation. To solve this problem, we make two improvements: (1) We use the multiple hops mechanism \cite{sukhbaatar2015end} in Dynamic Subtitle Memory to enhance the reasoning ability of the module. (2) We update the Dynamic Subtitle Memory to remove the irrelative information. From the Eq. \eqref{clip-level-ave-rep}, we can get the clip-level semantic representation $v$ by the Dynamic Subtitle Memory. Then we use ${v}$ to update the Dynamic Subtitle Memory. The framework is shown in Fig. \ref{fig:extend_2}. And the update procedures are computed as follows:
\begin{align}
  \gamma_n^t &= {ReLU}({(v^t)}^Ts_n^t), \label{weight-of-frame-update}\\
  s_n^{t+1} &= \gamma_n^ts_n^t, \label{update-of-sentences}
\end{align}
The $v^t$ is clip-level representation obtained by sum the output $v_i$ of the $t$-th Dynamic Subtitle Memory. And the $s_n^t$ is the ${n}$-th sentences of the $t$-th Dynamic Subtitle Memory. The ${\gamma_n^t}$ denotes the relationship between $v^t$ and $s_n^t$. We use the ReLU activate function to forget the irrelative memory and update the relative memory. The process of updating the frame-level and clip-level representation can be computed as follows:
\begin{align}
  \beta_i^n &\coloneqq {(v_i^t)}^Ts_n^{t+1}, \label{weight-of-frame-updated}\\
  v_i^{t+1} &= \sum_{n=1}^{N} \beta_i^ns_n^{t+1}, \label{weighted-sum-of-frames-extend}\\
  v^{t+1} &= \sum_{i=1}^{T}v_i^{t+1},\label{weighted-sum-of}
\end{align}
Note that, the `Dynamic' here has two meanings: (1) As described in the basic LMN, the Subtitle Memory is different for each movie. (2) In the Dynamic Subtitle Memory module with update mechanism, the Subtitle Memory will be updated by the clip-level representation.

\subsubsection{The Question-Guided Model.} We want to use the clip-level representation which is the output of the Dynamic Subtitle Memory module to answer questions. But the question information is not used in the Dynamic Subtitle Memory. So there will contain some information that can represent the clip-level movie content but is not relative to the question. Different from SMem-VQA \cite{xu2016ask} module which utilizes questions to attend to relevant regions, we propose a question-guided model to attend to the subtitles. We first utilize the questions to update the subtitles. Then the question-guided subtitles are applied to represent the movie clips. The question-guided subtitles can be computed by:
\begin{align}\label{question-guided-module}
  q_n &= softmax(u^Ts_n),\\
  s_n &\coloneqq q_ns_n,
\end{align}
where $Softmax(z_p)=e^{z_p}/\sum_{q}{e^{z_q}}$ and $u$ denotes the question representation. Suppose that both the subtitles and the question are embedded by Static Word Memory $W_e$ and followed by a mean-pooling over all words of the sentence to get the final representation. Thus we can obtain a question sensitive subtitles according to the similarity of each subtitle and question representation.

Besides the three extend frameworks described above, we combine the update mechanism and the question-guided Model together. In detail, after we get the final Dynamic Subtitle Memory which is updated by Eq. \eqref{update-of-sentences}, we use the question-guided model to update it again. Note that all the extended frameworks will not increase any learning parameters and thus are efficient.

\section{Experiments}\label{Experiments}
\subsection{Experimental Setup}
We evaluate the Layered Memory Network on the MovieQA dataset \cite{tapaswi2016movieqa}, which contains multiple sources of information such as video clips, plots, subtitles, and scripts. This dataset consists of 14,944 multiple-choice questions about 408 movies. Each question has five open-ended choices but only one of them is correct. All of them have subtitles but only 140 movies have video clips. We focus on the task of `Video+Subtitels'. The 6,462 questions-answer pairs are split into 4,318, 886, and 1,258 for training, validation, and test set, respectively. Also, the 140 movies (totally 6,771 clips) are split into 4385, 1098 and 1288 clips for training, validation, and test set, respectively. Note that one question may be relevant to several movie clips. As the test set only can be tested once per 72 hours on an online evaluation server. Following \cite{tapaswi2016movieqa}, we also split the training set (4,318 question-answer pairs) into 90\% train / 10\% development (dev.). As the answer type is multiple choices, the performance is measured by accuracy.

For video frame feature extraction, we first extract frames from all movie clips with the rate of 1 frame per second. Following \cite{tapaswi2016movieqa}, we also sample frames from all clips but only obtain $T=32$ movie frames with equal space. In our experiments, the extracted movie frames are first resized into $224 \times 224$. Then we extract features of `pool5' and `inception\_5b/output' layers from VGG-16 \cite{simonyan2014very} and GoogLeNet \cite{szegedy2015going}, respectively. The `pool5' layer has the shape of $512 \times 7 \times7$ while the `inception\_5b/output' layer has the shape of $1024 \times 7 \times 7$.

For the Static Word Memory, we utilize the word2vec model supported by \cite{tapaswi2016movieqa}, which is trained by skip-gram model \cite{mikolov2013efficient} on about 1200 movie plots. The word2vec model has the embedding dimension of 300. For fair comparisons and following \cite{tapaswi2016movieqa}, we also fix our Static Word Memory while training the models. For the Dynamic Subtitle Memory, we utilize all the sentences in each movie subtitle.

For training our LMN model, all the model parameters are optimized by minimizing the cross-entropy loss using stochastic gradient descent. The batch size is set to 8 and the learning rate is set to 0.01. We perform early stopping on the dev set (10\% of the training set).

\subsection{Experimental Results}

\begin{table}
 \small
 \caption{Accuracy for the proposed LMN model and two baseline models on the MovieQA dataset.  `V' and `G' denote VGG-16 and GoogLeNet, respectively.}\label{evaluate LMN model}
  \centering
  \begin{center}
  \begin{tabular}{l|c|c|c}
    \hline
    Model                                       & Video & Subtitles & Video+Subtitles \\
    \hline
    SSCB    & 21.6  & 22.3 & 21.9 \\
    MemN2N & 23.1  & 38.0 & 34.2 \\
    \hline
    LMN + (G)                                 & 38.3  & 38.1    & 39.3    \\
    \textbf{LMN + (V)}                                 & \textbf{38.6}  & \textbf{38.1}    & \textbf{39.6}    \\
    \hline
  \end{tabular}
  \end{center}

\end{table}

\subsubsection{Evaluate the Performance of LMN Model.} In this subsection, we evaluate the performance of our proposed Layered Memory Network (LMN). We compare LMN with two baseline models. The SSCB \cite{tapaswi2016movieqa} utilizes a neural network to learn a factorization of the question and answer similarity. The MemN2N model \cite{sukhbaatar2015end} is first proposed for text question answering (QA) and modified by \cite{tapaswi2016movieqa} for movie QA. All the results have been illustrated in Table \ref{evaluate LMN model}.

\begin{figure*}
  \centering
  \includegraphics[width=0.95\linewidth]{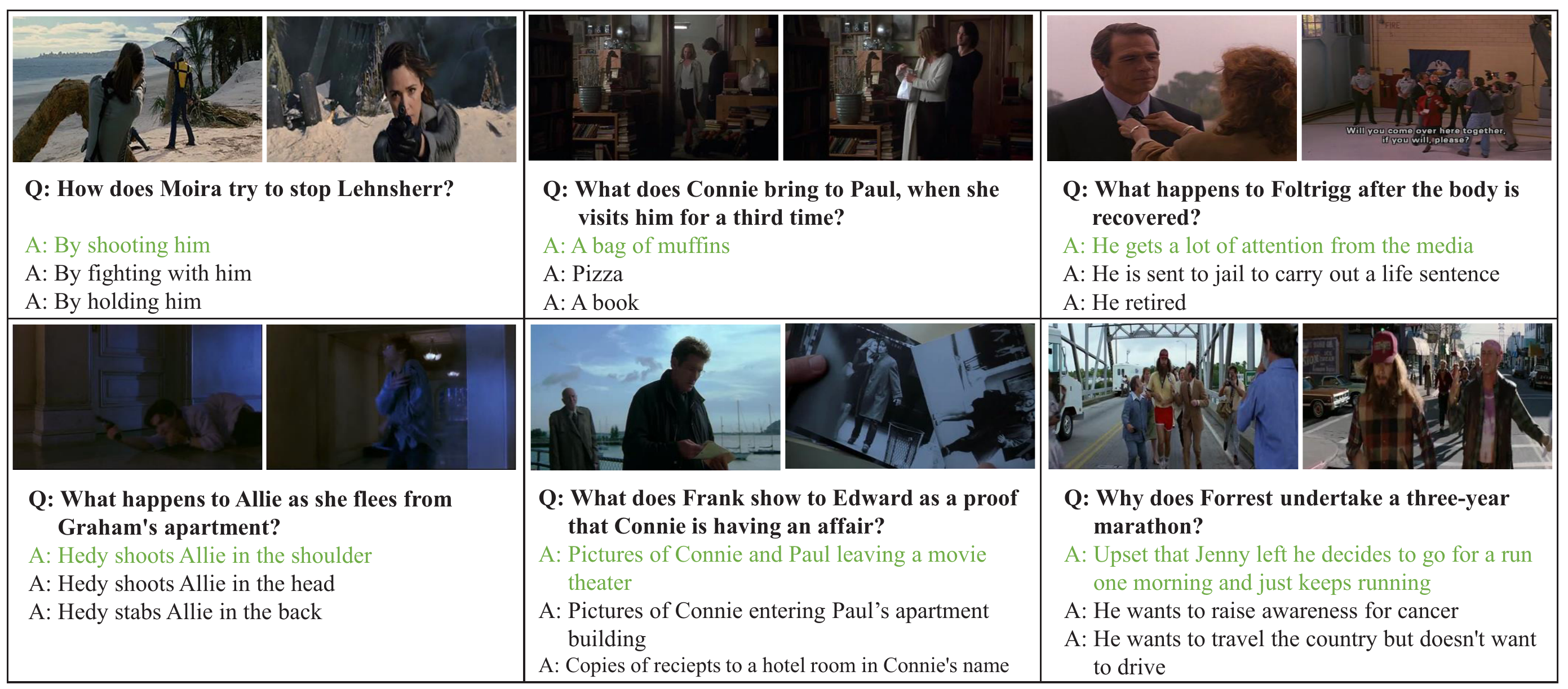}
  \caption{Examples of movie question answering from LMN with only frame-level representation. The correct answers (in green color) are predicted and hitted by LMN. We also list two more answer choices annotated in MovieQA for comparison.}\label{fig:example_1}
\end{figure*}

\begin{figure*}[ht]
\centering
\includegraphics[width=0.95\linewidth]{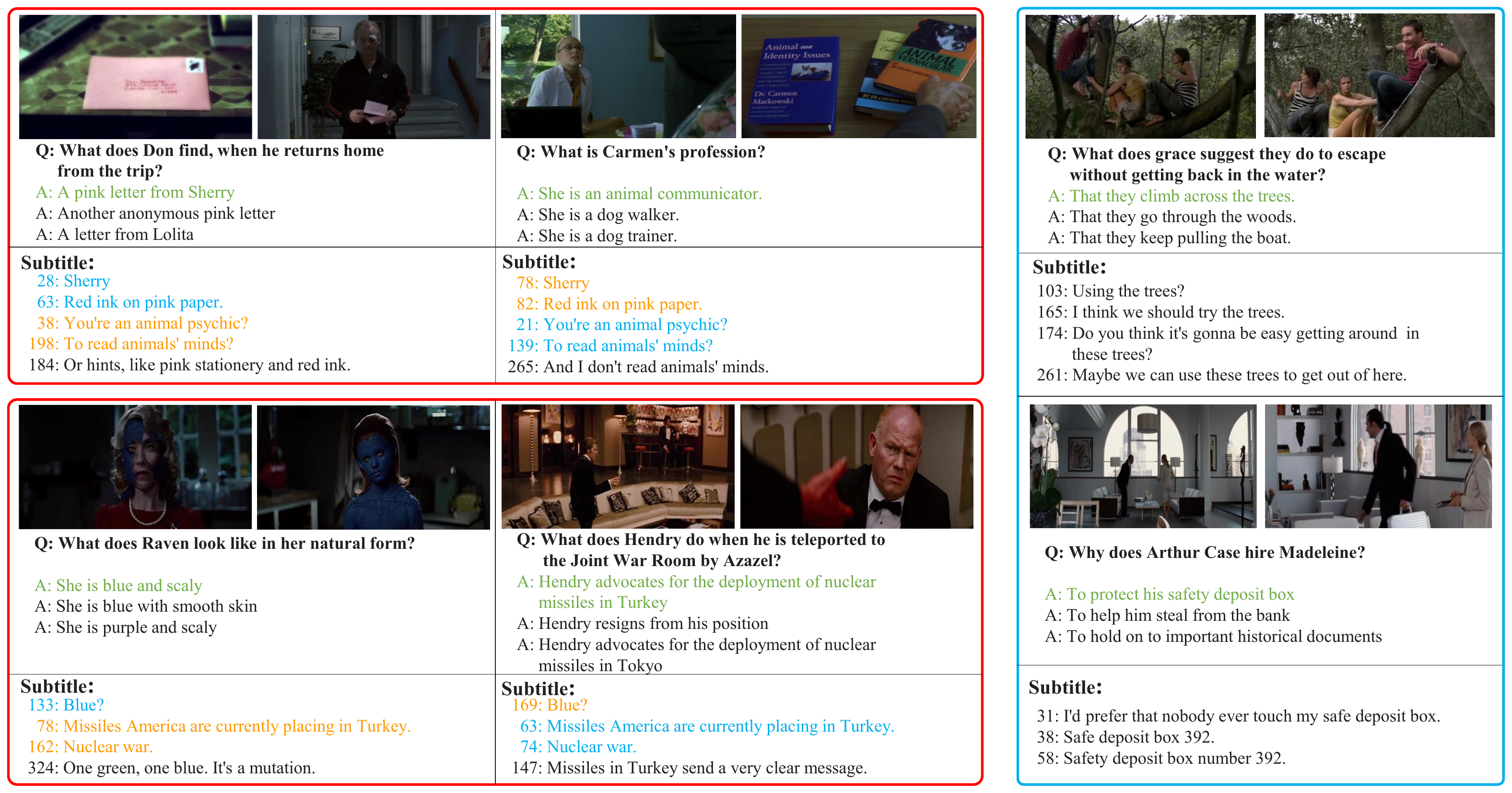}
\caption{Examples of movie question answering from LMN. For each clip, we also list serveral relevant subtitles. The number at the beginning of each line of subtitles is the rank index obtained from LMN according to the similarity between the frame-level representation and subtitle embedding, the smaller the more similar. Examples inside red bounding boxes are from the same movies. Subtitles in blue color are relevant to the questions, whereas the red ones are irrelevant.}\label{fig:example_2}
\end{figure*}

\begin{table}[ht]
  \small
  \caption{Accuracy for the extended frameworks on the MovieQA dataset. `V' and `G' denote VGG-16 and GoogLeNet, respectively. SWM represent the Static Word Memory. UM represent the update mechanism of dynamic subtitle memory. QG represent the question-guided model. And the superscript represents the number of SWM or UM.}\label{evaluate extended frameworks}
  \centering
  \begin{center}
  \begin{tabular}{l|c}
    \hline
    Model                                       &  Video+Subtitles\\
    \hline
    LMN + (G)                                        & 39.3   \\
    LMN + (V)                                        & 39.6  \\
    \hline
    LMN + Multiple hops in SWM$^2$ + (G)                    & 39.6  \\
    LMN + Multiple hops in SWM$^2$ + (V)                    & 39.6   \\
    LMN + Multiple hops in SWM$^3$ + (G)                    & 40.1  \\
    LMN + Multiple hops in SWM$^3$ + (V)                    & 39.9  \\
    \hline
    LMN + UM$^2$ + (G)                    & 41.4   \\
    LMN + UM$^2$ + (V)                    & 41.6   \\
    LMN + UM$^3$ + (G)                    & 41.5   \\
    LMN + UM$^3$ + (V)                    & 41.4   \\
    \hline
    LMN + QG + (G)         & 40.6    \\
    LMN + QG + (V)         & 40.1    \\
    \hline
    LMN + UM$^2$ + QG + (G)                 & 42.3 \\
    \textbf{LMN + UM$^2$ + QG + (V)}                 &\textbf{42.5} \\
    \hline
  \end{tabular}
  \end{center}
\end{table}

From Table \ref{evaluate LMN model}, we can first see that understanding movie stories by only exploiting video content is a really hard problem which can be proved by the near random-guess result of `SSCB' and `MemN2N' methods. Secondly, our LMN model with VGG-16 features obtains a large margin performance gains of 15.5\% by only exploiting the video content. Even compared with `MemN2N', which takes both video and subtitles as inputs, LMN still outperforms it by 4.4\%. Note that LMN only contains frame-level representations without exploiting movie subtitles. While on `Video+Subtitles' task, the `SSCB' has a near random-guess performance while `MemN2N' degrades the performance about 3.8\%. LMN obtains a further performance improvement of 1\%. We repeat the MemN2N model and obtain a compete performance of 37.45\%. This can illustrate that the performance improvement results from the effectiveness of LMN model. Finally, while using GoogLeNet features, our method obtains similar performance improvement. In summary, we can conclude that the semantic information (e.g., movie subtitles) is important for movie question answering and LMN can perform well on movie stories understanding even without subtitles.

\subsubsection{Evaluate the Performance of the Extended Frameworks.}
All the results have been shown in Table \ref{evaluate extended frameworks}. From the second block of Table \ref{evaluate extended frameworks}, we can see that the different number of Static Word Memory have the similar performance. The reasons might be: Firstly, the Static Word Memory is composed by a vocabulary size 26,630 and $d$ = 300. Although we use the multiple hops mechanism, it is difficult to get a more precise representation from this large memory. Secondly, in the multiple hops process, we get the regional representation of each frame without knowing anything about the question. we get the region representation but it may have no connection with the question and answers.

The third block of Table \ref{evaluate extended frameworks} shows the result of the update mechanism of Dynamic Subtitle Memory. We obtain a performance improvement of 2.0\% when use two Dynamic Subtitle Memories. This can illustrate that through the update mechanism the model does remove some irrelevant information and could reach a better understanding ability.

From the forth block of Table \ref{evaluate extended frameworks}, we can observe that LMN  with Question Guided extension obtains a performance improvement of 1.0\% by taking VGG-16 features as inputs. Besides the two extended frameworks we also combine update mechanism and the question-guided model together. The results are listed in the fifth row of Table \ref{evaluate extended frameworks}. We obtain a performance improvement of 0.9\% than that of only using the update mechanism. Thus the question-guided mode could make the Dynamic Subtitle Memory more relevant to questions and LMN has a good extendable capability.

\subsubsection{Examples of Movie Question Answering.}
We first show examples from LMN with only frame-level representation, i.e., with no subtitles incorporated. From examples in Fig. \ref{fig:example_1} we can see that, though there are no direct connections between words in questions and answers, LMN successfully hits the correct answers. There examples illustrate that LMN could accurately associate specific language elements to particular video content.

In Fig. \ref{fig:example_2}, we show examples from LMN model, i.e., with subtitles incorporated. From couples of examples inside the red bounding boxes we can see that, given the same set of candidate subtitles within one movie, LMN successfully infers relevant relationship between subtitles and question-answer pairs. For example, `Sherry' and `Red ink on pink paper' are ranked higher for the answer `A pink letter from Sherry' than that for the answer `She is an animal communicator', i.e., with rank index of `28' and `63' vs. `78' and `82'. Examples in the blue bounding box also illustrate the effectiveness of our method.

\begin{table}[ht]
  \small
  \caption{Accuracy comparison on the test set of MovieQA online evaluation server. The results are screened on Sept. 10, 2017. `MN' denotes Memory Networks.}\label{performance on test set}
  \centering
  \begin{center}
  \begin{tabular}{l|c}
    \hline
    Model                                       &  Video+Subtitles \\
    \hline
    OVQAP                                & 23.61 \\
    Simple MLP                            & 24.09 \\
    LSTM + CNN                             & 23.45 \\
    LSTM + Discriminative CNN             & 24.32 \\
    VCFSM                                & 24.09 \\
    DEMN \cite{ijcai2017-280}            & 29.97 \\
    RWMN \cite{na2017movie}                                     & 36.25 \\
    \hline
    LMN + (V) + (Video only)                           & 34.34 \\
    \textbf{LMN + UM$^2$ + QG + (V)}                        & \textbf{39.03} \\
    \hline
  \end{tabular}
  \end{center}
\end{table}

\subsubsection{Test Set Online Evaluation.}
We evaluated the proposed models with the test set on the MovieQA online evaluation server. All the results are shown in Table \ref{performance on test set}. We also list some results with top performance on the `Leader Board'\footnote{\url{http://movieqa.cs.toronto.edu/leaderboard/#table-movie}}. From Table \ref{performance on test set}, we can see that LMN with only frame-level representation obtains a compete performance. Particularly, LMN with update mechanism and question-guided model outperforms other methods by about 2.78\% in `Video+Subtitles' task. Moreover, LMN with update mechanism and question-guided model ranked the first (till Sept. 10, 2017).

\section{Conclusions}\label{Conclusions}
In this paper, we propose a Layered Memory Network (LMN) for movie question answering. LMN learns a layered representation of movie content, which not only encodes the correspondence between words and visual content inside frames but also encodes the temporal alignment between sentences and frames inside movie clips. We also extend LMN model to three extended frameworks. Experimental results on the MovieQA dataset show the effectiveness and better performance of our method. We also illustrate the effectiveness by movie question answering examples. In addition, on the online evaluation server, LMN with update mechanism and question-guided together ranked the first on the `Video+Subtitles' task.

\section{Acknowledgments}
This work is supported by the NSFC (under Grant U1509206, 61722204, 61472116, 61472276).

\bibliographystyle{aaai}
\bibliography{sigproc}
\end{document}